# A Feature-Level Ensemble Model for COVID-19 Identification in CXR Images using Choquet Integral and Differential Evolution Optimization

**Amir Reza Takhsha[1], Maryam Rastgarpour[2] and Mozhgan Naderi[3]**

**Abstract**

The COVID-19 pandemic has profoundly impacted billions globally. It challenges public health and healthcare systems due to its rapid spread and severe respiratory effects. An effective strategy to mitigate the COVID-19 pandemic involves integrating testing to identify infected individuals. While RT-PCR is considered the gold standard for diagnosing COVID-19, it has some limitations such as the risk of false negatives. To address this problem, this paper introduces a novel Deep Learning Diagnosis System that integrates pre-trained Deep Convolutional Neural Networks (DCNNs) within an ensemble learning framework to achieve precise identification of COVID-19 cases from Chest X-ray (CXR) images. We combine feature vectors from the final hidden layers of pre-trained DCNNs using the Choquet integral to capture interactions between different DCNNs that a linear approach cannot. We employed Sugeno-$\lambda$ measure theory to derive fuzzy measures for subsets of networks to enable aggregation. We utilized Differential Evolution to estimate fuzzy densities. We developed a TensorFlow-based layer for Choquet operation to facilitate efficient aggregation, due to the intricacies involved in aggregating feature vectors. Experimental results on the COVIDx dataset show that our ensemble model achieved 98% accuracy in three-class classification and 99.50% in binary classification, outperforming its components—DenseNet-201 (97% for three-class, 98.75% for binary), Inception-v3 (96.25% for three-class, 98.50% for binary), and Xception (94.50% for three-class, 98% for binary)—and surpassing many previous methods.

## 1. Introduction

COVID-19, caused by the SARS-CoV-2 virus, was first found in Wuhan in December 2019 [1]. It was declared a global pandemic in March 2020 [2] and still remains a problem, even though the WHO no longer calls it a public health emergency [3]. Its ability to spread easily, even among people without symptoms, makes it hard to control [4]. By March 2024, there were over 704 million reported cases worldwide [5]. Effective vaccines play a key role in ending the pandemic [6]; however, despite their availability, achieving complete eradication of COVID-19 remains impossible [7]. Since the pandemic began, variants like Alpha, Beta, Delta, and Omicron have emerged [8]. Omicron, being more transmissible, poses challenges by reducing vaccine efficacy and speeding up immunity loss [7]. Even after four years, COVID-19 continues to spread globally, driven by fast-spreading Omicron subvariants [9].

Moreover, there is concern about a new variant that could avoid the protection of current COVID-19 vaccines [10]. A stronger variant would put more strain on health systems, showing we are not fully prepared [11]. Because of this, we must focus on being ready to deal with the changing COVID-19 situation.

An effective COVID-19 pandemic response strategy includes integrating testing to detect COVID-19 patients, isolation, and contact tracing [12]. The Reverse Transcriptase-Polymerase Chain Reaction (RT-PCR) test is widely recognized as the gold standard for COVID-19 diagnosis. However, the RT-PCR testing method for COVID-19 diagnosis can be time-consuming [13] and prone to generating false negative results due to sampling and testing errors [14].

Computer-Aided Detection and Diagnosis (CAD) systems can reduce false negative rates and address errors in perception [15]. CAD systems are divided into two types: Computer-Aided Detection (CADe), which identifies abnormalities in medical images, and Computer-Aided Diagnosis (CADx), which characterizes lesions [16]. These systems are designed to provide diagnostic results without human subjectivity [17]. They also help alleviate the burden on medical care systems [18]. As a result, CAD systems can address challenges such as the strain on the healthcare system in China. This strain occurred due to the surge in Omicron variant infections, despite extensive vaccination drives [19]. Moreover, uncommon medical resources and overwhelmed healthcare systems in resource-limited areas pose challenges for COVID-19 control [20]. Given that symptoms overlap with Pneumonia [21], expert radiologists are essential for accurate disease identification. Therefore, an effective CAD system that detects lung lesions in resource-limited areas is crucial for stopping the spread of COVID-19 and ensuring the best patient care.

---

[1] Faculty of Engineering, Department of Computer Engineering, Islamic Azad University, Saveh, Iran
  amirrezatakhsha@gmail.com, amirreza.takhsha@iau.ac.ir

[2] Faculty of Engineering, Department of Computer Engineering, Islamic Azad University, Saveh, Iran
  m.rastgarpour@iau-saveh.ac.ir; m.rastgarpour@gmail.com

[3] Faculty of Engineering, Department of Computer Engineering, Islamic Azad University, Saveh, Iran
  mozhganaderii@gmail.com, m.naderi@stu.iau-saveh.ac.ir

Deep learning-based CAD systems showcase potential in disease identification through radiological imaging and excel in detecting COVID-19 from medical images, as evidenced by various studies [22-25]. In these studies, Deep Convolutional Neural Networks (DCNNs) are used for detecting COVID-19 from Chest X-ray (CXR) images. DCNNs have the ability to learn complex representations and are the most commonly used type of deep learning-based models [26]. These networks efficiently demonstrate the capability to detect COVID-19. Unlike RT-PCR tests, CXR imaging is readily available in hospitals, providing affordable and prompt image acquisition [27]. Additionally, X-ray imaging is a cost-effective alternative to Computed Tomography (CT) scanning, contributing to its widespread usage during the pandemic [28]. Therefore, a DCNN-based diagnostic system for identifying COVID-19 from CXR images can provide a cost-effective, timely, and accurate approach for effective pandemic response efforts.

In deep learning, combining multiple models with an aggregation function creates a single model that benefits from different architectures. This approach, known as ensemble learning, addresses overfitting by harnessing the diversity of models to construct a more powerful model than its individual components [29]. In ensemble learning, the most commonly used aggregation function for combining models is the unweighted average, which assigns equal weight to all models in the ensemble. In contrast, the weighted sum aggregation assigns importance-based weights to each model, resulting in improved prediction accuracy compared to the unweighted average [29]. However, similar to the unweighted average, the weighted sum aggregation overlooks the interactions between models within the ensemble.

Conventional additive measures alone fail to adequately address interactions and mutual influences among attributes or criteria in real-world systems. In these scenarios, criteria often show interdependent or interactive behavior. This behavior contradicts the additive nature of these functions [30]. In ensemble deep learning, interdependencies may arise from shared training data and shared architectural components, and the degree of independence among models remains uncertain. Unlike linear ensemble operators, fuzzy integrals such as the Choquet integral and Sugeno integral can manage information aggregation under different assumptions about the independence of information sources. They are capable of handling situations where independence is either present or uncertain.

The Choquet Integral generalizes the weighted sum by using fuzzy measures [31]. Due to this generalization, it is applicable in fields like multivariate analysis, decision-making, pattern recognition, image and speech processing, and expert systems [32]. Indeed, fuzzy measures, which are a nonadditive set function [32], replace the weighted vector in the weighted sum [33]. This allows for the representation of the importance of both individual criteria and any combination of them [34]. The utilization of weights within the weighted sum can be interpreted as a specific manifestation of an importance measure in situations where criteria do not interact [32]. In contrast, fuzzy measures, which can model subsets of criteria, allow for the representation of both redundancy and complementarity among criteria [31,35]. These factors are important considerations in real-world applications.

Before applying aggregation with the Choquet integral, establishing fuzzy measures is essential. However, computing fuzzy measures in real-world scenarios is a complex task [35]. The complexity lies in the need to collect evaluations for $2^n - 2$ subsets, where the universal set comprises N criteria, in order to establish fuzzy measures. Addressing this complexity, our study employs a specific class of fuzzy measures called Sugeno λ-measures, in conjunction with the Differential Evolution algorithm, to compute fuzzy measures. We utilized Differential Evolution to calculate fuzzy measures for singletons or individual DCNNs while utilizing Sugeno λ-measures to compute fuzzy measures for subsets of DCNNs.

In this study, we aggregate feature vectors from the Multilayer Perceptron (MLP) sections of different DCNN models. This approach goes beyond simple vote aggregation using linear operators by better capturing complex patterns through the Choquet integral and nonadditive fuzzy measures. These measures incorporate diverse perspectives from feature vectors in the higher layers of various DCNNs with different architectures.

The key contributions of our paper are outlined as follows:

In our research, we proposed a DCNN-based framework that leverages ensemble learning to integrate models with diverse architectures. Unlike previous approaches, this framework utilizes the Choquet integral to aggregate the flattened feature vectors from pre-trained DCNNs-specifically DenseNet-201, Inception-v3, and Xception-to enhance the detection of COVID-19 cases from Pneumonia and Normal cases.

We developed a parallel-accelerated Choquet layer within the TensorFlow framework to enable the efficient combination of multiple sets of feature vectors, which have higher dimensionality compared to confidence score vectors, requiring an efficient algorithm to manage the increased complexity.

We addressed the complexity of computing fuzzy measures by incorporating a different approach into our proposed method. We used the Differential Evolution optimization algorithm to compute near-optimal fuzzy densities for each DCNN. Next, we applied Sugeno-λ measures to determine the fuzzy measures for subsets of DCNNs.

The rest of the paper is structured as follows: Section 2 reviews related works. Section 3 provides a comprehensive description of our proposed method; Section 4 presents the experimental results obtained from the COVIDx Dataset; Section 5 conducts a comparative analysis of our proposed ensemble model's results with previous methods and Section 6 presents the conclusion and outlines future directions.

## 2. Related Work

Amid the challenges presented by COVID-19, recent advancements in technology, particularly in areas like deep learning, offer significant promise for improving global healthcare. With CXR imaging playing a vital role in cost-effective and prompt COVID-19 detection, this section underscores the latest developments in employing deep learning models to identify the virus from CXR images.

Wang et al. [24] introduced COVID-Net, a customized DCNN model designed through a human-machine collaborative strategy to detect COVID-19 from CXR images. Using a machine-driven approach guided by expert input, they optimized macro and micro architecture designs, including layer arrangements and activation functions. This led to a lightweight PEPx design pattern with two 1×1 convolutions, a 3×3 depth-wise convolution, and two more 1×1 convolutions. COVID-Net includes 16 PEPx blocks and uses MLP for classification. After being trained on the COVIDx dataset, the model achieved 93.3% accuracy in a three-class classification task (COVID-19, Pneumonia, Normal).

Ucar and Korkmaz [36] introduced COVIDdiagnosis-Net, a compact deep neural network designed for rapid COVID-19 detection. Built on Deep Bayes-SqueezeNet, a fusion of SqueezeNet and Bayesian optimization, this model benefits from SqueezeNet's smaller size, enabling efficient hardware deployment critical for combating COVID-19. To address the imbalanced COVIDx dataset, Offline Augmentation was used, boosting accuracy to 98.83% in three-class classification. Without data augmentation, the model's accuracy was 76.37%.

Hiedari et al. [37] utilized the VGG-16 model to detect COVID-19 in CXR images. They applied transfer learning to acquire the model's weights, mitigating the risk of overfitting linked to training on a dataset with limited COVID-19 cases. After fine-tuning, the model achieved an accuracy of 94.5% for three-class classification (Normal, Pneumonia, and COVID-19) and 98.1% for two-class classification (COVID-19 and non-COVID-19).

Similarly, Minae et al. [25] and Hemdan et al. [38] employed various pre-trained models for the task of COVID-19 detection. Minae et al. [33] employed pre-trained ResNet-18, ResNet-50, SqueezeNet, and DenseNet-169, and fine-tuned the last layer of the pre-trained versions of these DCNNs for COVID-19 and non-COVID-19 classification. The best performer, SqueezeNet, achieved a sensitivity rate of 98% and a specificity rate of 92.9%. Hemdan et al. [38] utilized Inception-v3, VGG-19, DenseNet-201, Xception, MobileNetv2, and Inception-ResNet-V2 for two-class classification (COVID-19 and Normal). Among these models, VGG-19 and DenseNet-201 proved most successful, both achieving a precision of 83% for the COVID-19 class.

Turkoglo [39] used the pre-trained AlexNet to extract features from CXR images for COVID-19 diagnosis. The Relief algorithm, based on KNN principles, identified and prioritized key features across all layers. These selected features were input into an SVM classifier, achieving 99.2% accuracy in a three-class task (Normal, COVID-19, Pneumonia) using 1500 deep features.

Chandra et al. [40] introduced the Automatic COVID Screening (ACoS) system, which uses tissue descriptors from CXR images to distinguish between normal, abnormal, COVID-19, and Pneumonia cases. The system extracts feature vectors such as first-order statistics, gray-level co-occurrence matrix, and histograms of oriented gradients. The Binary Grey Wolf Optimization (BGWO) algorithm selects relevant features for training five supervised models: Decision Tree (DT), SVM, KNN, Naïve Bayes (NB), and Artificial Neural Network (ANN), chosen for their ability to perform well with small datasets. An ensemble majority voting strategy is used to reduce misclassification, particularly false negatives. ACoS achieved 91.329% accuracy in classifying COVID-19 from Pneumonia CXR images.

Bhowal et al. [23] proposed a fuzzy Choquet ensemble model using pre-trained VGG-16, Xception, and Inception-V3 models for feature extraction, followed by input into an MLP component. Fuzzy measures were computed using coalition game theory, information theory, and Sugeno-λ measures, with the Shapley value used to evaluate individual classifier contributions. Three weighting schemes based on validation accuracies were applied, leading to the calculation of three fuzzy measures. These measures were then used to compute Choquet integrals, which were aggregated through majority voting. The model achieved 93.81% accuracy in a three-class classification on the COVIDX dataset.

Dey et al. [41] implemented a fuzzy Choquet integral ensemble method to combine votes from pre-trained VGG19, Inception-Netv3, and DenseNet-121. Using Sugeno-λ measures, they calculated fuzzy membership values for each DCNN by dividing its validation accuracy by the sum of all DCNNs' validation accuracies. The model achieved 99.02% accuracy in a three-class classification (COVID-19, Normal, Pneumonia).

Banerjee et al. [22] introduced COFE-NET, where they employed the Choquet integral to integrate the outputs of three pre-trained DCNNs—Inception-V3, DenseNet-201, and InceptionResNetv2—fine-tuned for the task of COVID-19 detection using CXR and CT images. Sugeno-λ measures theory was applied to determine fuzzy measures for subsets of the DCNNs. They adopted a trial-and-error approach to determine the fuzzy membership value of each DCNN. COFE-NET attained a 96.39% accuracy for three-class classification and a 99.49% accuracy for binary class classification on the COVIDX dataset.

Ukwuoma et al. [42] developed a deep learning framework for COVID-19 detection by integrating pre-trained DenseNet-201, VGG-16, and Inception-V3 models for feature extraction. The features from these models were concatenated after applying zero-padding to their final convolutional layer outputs. A Multi-head Self-attention network processed the concatenated features, addressing important details missed by max pooling layers. An MLP classification layer with a softmax activation function was used for prediction. The model achieved 96.33% accuracy in multi-class classification (COVID-19, Pneumonia, Lung Opacity, Normal) and 98.67% accuracy in binary classification (COVID-19, Normal).

Ulah et al. [43] proposed MTSS-AAE, a Multi-Task Semi-Supervised learning framework for COVID-19 detection using Adversarial Autoencoders (AAE). AAE combines autoencoder principles with adversarial training to improve the robustness of learned features. In the unsupervised phase, they initialized the model using the CheXpert [44] dataset, helping the network capture relevant data representations. In the supervised phase, the encoder learns to generate latent representations, enabling the classifier to detect COVID-19 with limited data. The semi-supervised phase incorporates additional tasks such as Pneumonia and Lung Opacity. The model achieved 96.95% accuracy on the COVIDx dataset, with improved performance from the auxiliary tasks.

Zhang et al. [45] proposed CXR-Net, a multi-task deep learning model for COVID-19 Pneumonia diagnosis using CXR images. It combines feature extraction, image reconstruction, and explainability, generating high-resolution visual explanations. CXR-Net achieved 87.9% accuracy in classifying Bacterial Pneumonia, Viral Pneumonia (COVID-19 and non-COVID-19), and Healthy cases.

Nayak et al. [46] introduced LW-CORONet, a lightweight model for efficient COVID-19 detection from CXR images. The architecture includes three Convolution, Batch Normalization, and ReLU blocks, followed by a flatten layer, Batch and Dropout block, and fully connected layers with ReLU activation. LW-CORONet reduces computational costs, minimizes parameters, and accelerates learning. It achieved 96.25% accuracy in a three-class classification task using the COVIDx dataset.

Akyol [47] proposed the ETSVF-COVID19 model for COVID-19 diagnosis using transformer-based frameworks applied to CXR and CT images. The model leverages advanced transformer architectures (ViT, Swin Transformer, LeViT, PoolFormer, and SwinV2) to extract deep features from medical images. These features are classified using machine learning models such as MLP, SVM, RF, and XGBoost. The framework employs a two-stage majority voting process: first, combining predictions from different classifiers, and second, further combining the results from the first stage. ETSVF-COVID19 achieved high accuracy, 99.2% for CT images and 98.56% for CXR images, in classifying Normal, COVID-19, and Pneumonia cases.

Salama et al. [48] proposed a hybrid model for COVID-19 detection using chest CT images. They employed 10 different DCNNs — Xception, Darknet53, VGG19, AlexNet, GoogLeNet, MobileNetV2, SqueezeNet, Darknet19, ResNet50, and ResNet101 — for feature extraction and five machine learning classifiers — SVM, KNN, Ensemble Classifier, Naive Bayes, and DT — for classification. Their method identified the optimal CNN layers for feature extraction and the best classifiers for categorizing images into COVID-19 and non-COVID-19 cases. The highest accuracy rate of 99.39% was achieved using MobileNetV2's layer 84 as the feature extractor with the SVM classifier.

## 3. Proposed Method
*3.1. Method Overview*

We developed an ensemble model using pre-trained DCNNs, specifically DenseNet-201 [49], Inception-v3 [50], and Xception [51], as base models. Each DCNN was modified with an MLP classification section for both three-class and two-class classification tasks. The MLP includes a flattening layer, a dense layer with 100 ReLU-activated neurons, a dropout layer for regularization, another dense layer with 256 ReLU-activated neurons, and a prediction layer with either three or two neurons depending on the classification task. The softmax function is used in the prediction layer to generate a probability distribution for each CXR image.

Instead of using vote aggregation, we integrate the outputs of the dense layers (256 ReLU-activated neurons)

from all three DCNNs by feeding these outputs into a Choquet layer. After passing through the MLP blocks, each DCNN generates a feature vector of shape (Batch Size, 256). These vectors are then passed into the Choquet layer, which aggregates the activations into a unified tensor of shape (Batch Size, 256). Finally, this tensor is input into the prediction layer of DenseNet-201 to produce the final likelihood vector for each image, completing the ensemble model's construction. Determining the fuzzy measures required by the Choquet layer for effective aggregation is a complex task. To streamline the process of identifying the best-fit Choquet parameters, we compute fuzzy measures for subsets of DCNNs based on Sugeno-λ measure theory. Consequently, fuzzy measures for subsets of DCNNs can be derived by first determining the fuzzy measures for singletons. In this study, the fuzzy measures for singletons are calculated using a meta-heuristic algorithm—Differential Evolution.

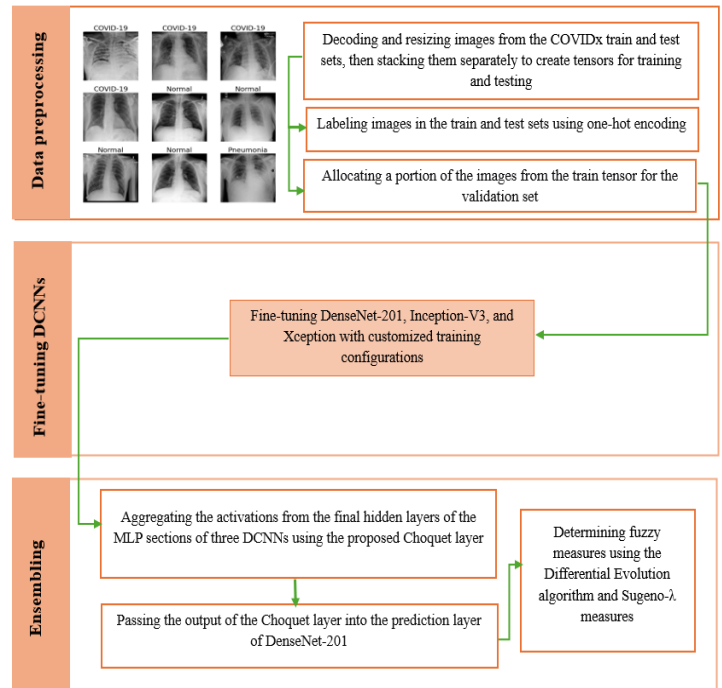

**Fig. 1** Block diagram illustrating our proposed model for identifying COVID-19 from CXR images.

### 3.2. Data Preprocessing

In the preprocessing stage, images from both the training and testing datasets of the COVIDx dataset were read and decoded into pixel data, which represents color information in the RGB (Red, Green, Blue) color space. Each image was then converted into a three-dimensional tensor, where the first dimension corresponded to the image's height, the second to its width, and the third to the color channels (Red, Green, Blue). To meet the input requirements of pre-trained DCNNs, all tensors were resized to a standardized dimension of 224×224×3. This resizing was accomplished using the Bilinear Interpolation algorithm. The algorithm efficiently computes new pixel values. It does this by averaging neighboring pixels from the original image. This process is efficient and not computationally intensive. To facilitate batch mode training and testing of DCNNs with four-dimensional input layers, the three-dimensional tensors are stacked together, resulting in the creation of a four-dimensional tensor array. To ensure consistency and optimal readiness for subsequent adjustments of the DCNNs, pixel values were rescaled to a uniform range of [0, 1]. This was achieved by applying simplified min-max normalization, in which the values of the four-dimensional tensor were divided by 255. The rescaling step plays a crucial role in ensuring data uniformity and enhancing the readiness of the data for subsequent model refinement procedures.

### 3.3. Transfer Learning

We employed the transfer learning technique to counter the limited COVID-19 images in the COVIDx dataset and alleviate overfitting risks. Pre-trained weights from models trained on the ImageNet dataset [52], which includes diverse images such as Zucchini flower, Ostrich, Lemur, Quokka, Gharial, Mandrill, and Okapi, were loaded into three DCNN models. Although there are differences between ImageNet and medical datasets like COVIDx, transferring knowledge from ImageNet helps DCNNs capture low-level features. This transfer is effective, even though the object classes differ. In our transfer learning approach, we employed a fine-tuning technique to optimize the weights of layers in the DCNNs specifically for the CXR dataset.

### 3.4. DCNNs and MLP Blocks

The effectiveness of ensemble models comes from combining networks with different approaches to feature extraction. DenseNet-201, Inception-V3, and Xception, each built on distinct architectures, introduce valuable diversity into the ensemble. DenseNet-201 offers densely connected feature maps, Inception-V3 focuses on multi-scale extraction, and Xception uses depthwise-separable convolutions for efficient representation. Incorporating this diversity and combining these diverse architectures enriches feature representation and boosts performance. In this section, we briefly describe the different and diverse architectures of the three DCNNs, followed by the configurations of their MLP blocks.

#### 3.4.1. DenseNet-201

DenseNet-201, based on the DenseNet architecture introduced by Huang et al. [49], is known for its dense blocks and connectivity. Each Dense Block consists of multiple layers, where each layer performs batch normalization, ReLU activation, and a 3×3 convolution. Dense connectivity ensures direct connections from each layer to all subsequent layers, improving information flow, feature reuse, and gradient flow, which enhances model performance.

Furthermore, DenseNet incorporates a bottleneck structure by utilizing a 1×1 convolution before each 3×3 convolution. In each dense layer that uses a bottleneck structure, the process starts with a batch normalization operation, followed by a ReLU activation. This is succeeded by a 1×1 Convolution operation. Subsequently, another round of batch normalization operation and ReLU activation layer follows, leading to a final 3×3 convolution operation. The bottleneck layers compress feature maps and decrease the number of input channels before transitioning into more computationally intensive layers.

To manage feature map dimensions and the number of channels as the network progresses, DenseNet employs transition layers between dense blocks. Each transition layer starts with a batch normalization layer. Subsequently, a 1×1 convolutional operation is applied, followed by a 2×2 average pooling layer to downsample the feature maps.

DenseNet consists of three models: DenseNet-201, DenseNet-169, and DenseNet-121, each named for the total number of layers they contain. For instance, in DenseNet-201, the number 201 indicates the total number of layers.

#### 3.4.2. Inception-V3

Inception-V2 and Inception-V3, developed by Szegedy et al. [50], build upon the Inception-V1 architecture, which was designed to address the challenge of handling images with features of varying sizes. The core innovation of Inception-V1 was the "Inception module," which uses convolutional layers with different filter sizes (1×1, 3×3, and 5×5) alongside a pooling layer. These layers are combined through concatenation to improve efficiency. To optimize computation, 1×1 convolutional layers are placed before and after the larger convolution layers and pooling layers to reduce input channels. Additionally, Inception-V1 introduced auxiliary classifiers to improve gradient flow in deep networks, helping mitigate issues during backpropagation.

Inception-V2 enhances the basic Inception module by factorizing the 5×5 convolution into two consecutive 3×3 convolutions, improving computational efficiency. This is because two 3×3 convolutions require fewer multiplications than a single 5×5 convolution. Additionally, Inception-V2 further optimizes convolutions by breaking down n×n convolutions into 1×n and n×1 convolutions within each Inception module. Rather than increasing the depth of the Inception module, Inception-V2 expands the filter bank, using filters of various sizes (both width and height) in parallel convolution layers. This strategy alleviates representational bottlenecks, enabling the network to learn a broader range of features while reducing computational costs and model complexity.

To enhance the Inception-V2 architecture without significant modifications to its modules, the authors introduced Inception-V3. This version incorporates several improvements, including the utilization of the RMSProp optimizer and Factorized 7×7 convolutions. Additionally, batch normalization is applied in the auxiliary classifiers, and label smoothing regularization is implemented to prevent the model from becoming excessively confident in its predictions.

#### 3.4.3. Xception

Xception, proposed by Chollet et al. [51], introduces a DCNN architecture that relies heavily on Depthwise separable convolution layers. The key idea behind Xception is to separate channel-wise and spatial dependencies within feature maps. This concept builds upon the Inception architecture. Depthwise separable convolutions achieve this separation through two steps: Depthwise convolution and pointwise convolution. In Depthwise convolution, each input channel is convolved with a separate filter, unlike traditional convolutions that apply a single filter to all channels. This allows the network to efficiently capture spatial information while reducing computational costs. The pointwise convolution (1×1) follows, transforming the output channels from the Depthwise convolution into a different dimensionality.

#### 3.4.4. MLP Blocks

The MLP block at the top of each DCNN uses a feedforward design. Data flows from the flattened input layer

(from the convolutional section) through hidden layers to the output layer without feedback. Each neuron connects to all neurons in the previous and next layers. The neuron output is calculated as a weighted sum of inputs, plus a bias, and passed through a ReLU activation function. The bias helps adjust decision boundaries, and ReLU adds non-linearities to improve pattern learning.

For each DCNN, after flattening the extracted features, the data passes through a dense layer with 100 neurons and ReLU activation, followed by dropout regularization. Each DCNN uses a specific dropout rate: 0.4 for DenseNet, 0.3 for Inception, and 0.2 for Xception. A dropout rate, like 0.4, means there is a 40% chance of deactivating each neuron during training. After the dropout layer, another dense layer with 256 neurons and ReLU activation serves as the final hidden layer. The output from this layer goes to the prediction layer, which has either three neurons (for three classes) or two neurons (for two classes). The outputs are processed with a softmax function to compute the probability distribution.

### 3.5. Training Parameters

We use Stochastic Gradient Descent (SGD) to fine-tune all DCNNs with the same settings: an initial learning rate of 0.001, a momentum of 0.9, and a clip-norm of 1.0. Categorical Cross Entropy serves as the loss function, and we monitor validation loss after each epoch. If the validation loss plateaus or increases after the 14th epoch, we reduce the learning rate by 40% to improve convergence. We save the model weights from the epoch with the highest validation accuracy as ".h5" files. After fine-tuning, we reload these weights into their respective models. Each DCNN undergoes 50 epochs of fine-tuning with a batch size of 32.

### 3.6. Proposed Fuzzy Choquet Ensemble Model

Given our preference for feature combination over vote aggregation, selecting an efficient aggregation method is crucial. Dense layers with 256 neurons have higher dimensionality than the prediction layers of individual DCNNs. This makes it essential to merge feature vectors rather than probabilities, especially when using functions like the Choquet integral. To achieve this, we use TensorFlow and Keras to implement a custom Choquet integral layer. This custom layer optimizes computation for both CPU and GPU platforms. It leverages TensorFlow's parallelism and computational efficiency. Using TensorFlow ensures compatibility with existing deep learning workflows. It also simplifies experimentation and deployment while reducing resource usage. With this setup, we route the outputs of the dense layers to the Choquet layer. The output of the Choquet layer is then fed directly into DenseNet-201's prediction layer.

To effectively utilize the Choquet integral for fusing information vectors, it's imperative to take into account fuzzy measures for subsets of DCNNs and fuzzy density associated with each DCNN. The coordination of fuzzy measures effectively manages interactions among Criteria (DCNNs) and leads to an enhanced scrutiny of relationships. However, obtaining all possible fuzzy measures for a given universal set requires human-collected evaluations for each subset, with the exclusion of the empty and universal sets. For additional clarity, our universal set consists of {DenseNet-201, Inception-V3, Xception}, thus necessitating the determination of optimal fuzzy measures for {Inception-V3}, {DenseNet-201}, {Xception}, {Inception-V3, Xception}, {DenseNet-201, Inception-V3}, and {DenseNet-201, Xception}, which is indeed a complex task. To streamline this, we employ Sugeno-λ measures proposed by Sugeno [53]. However, it remains essential to compute fuzzy measures for singletons (fuzzy densities) such as {Inception-V3}, {DenseNet-201}, and {Xception}. In this study, near-optimal fuzzy densities are calculated using the Differential Evolution algorithm, which explores a fuzzy space where the membership value of each element of the ensemble set ranges from 0 to 1.

#### 3.6.1 Fuzzy Measures Calculation using Sugeno-λ Measures Theory

A fuzzy measurement assigns a degree of membership to subsets of a given set. According to [53], a measurement $\Omega$ in a finite set X is considered a fuzzy measurement if its measurement function $g_\Omega: 2^x \to [0, \infty)$ satisfies the following conditions:

Nullity at the empty set and unity at the entire set: $g_\Omega(\Phi) = 0 \; and \; g_\Omega(X) = 1$.

Monotonicity: If $A \subseteq B$, then $g(A) \leq g(B)$.

Limit preservation: If $\{A_i\}_{i=1}^\infty$ is an ascending measurable sequence then $g_\Omega(A_i) = g_\Omega \left( \lim_{i \to \infty}(A_i) \right)$.

According to [53], the Sugeno- λ measures can be expressed as follows:
If A, B ⊆ X and $A \cap B = \emptyset$ then:

$$g_\Omega (A \cup B) = g_\Omega (A) + g_\Omega (B) + \lambda \, g_\Omega (A) \, g_\Omega(B) \text{ for some } \lambda > -1. \qquad (1)$$

The equation necessary to calculate the value of λ, as outlined in [54], can be derived using the following expression:

$$1 + \lambda = \prod_{i=1}^{N}(1 + \lambda \, g_\Omega^i) \qquad (2)$$

Here, $g_\Omega^1$, $g_\Omega^2$, and $g_\Omega^N$ represent singletons or fuzzy densities, which in this study are fuzzy measures of three DCNNs. By knowing the fuzzy densities, we can determine fuzzy measures for all combinations.

#### 3.6.2 Finding Fuzzy Densities

To calculate fuzzy densities and enable aggregation, this study uses the Differential Evolution algorithm. This technique, introduced by Storn et al. [55], is highly regarded as one of the top choices for solving intricate optimization challenges. It offers superior scalability compared to some existing Metaheuristic Search Algorithms (MSAs). This makes it well-suited for handling large-scale and computationally intensive optimization problems. The advantage comes from its reduced storage requirements. The four stages of the Differential Evolution algorithm—population initialization, mutation, crossover, and selection—are detailed in the following subsections for a clearer understanding of our proposed method.

### 3.6.2.1 Population

In the Differential Evolution, the population represents a set of candidate solutions utilized to explore and search the solution space. The population size, designated by NP and set to 15 in this study, defines the number of individuals in each generation, denoted as G. In this context $S_G = \{x_{j,G} \mid j = 1, 2, 3 \ldots NP\}$ represents the collection of individual solutions and $x_{j,G} = \{x_{1,j,G}, x_{2,j,G} \ldots x_{D,j,G}\}$ signifies a D-dimensional vector, denoting an individual solution. Each element of $S_G$ represents a set of candidate fuzzy densities for DCNNs, and each element within $x_{j,G}$ represents a candidate fuzzy density for a unique DCNN. Since we aim to find near-optimal fuzzy densities for three DCNNs, our problem is three-dimensional, with D equaling 3. Initially, without any near-optimal fuzzy densities to consider as solutions, candidate solutions are formulated. These solutions adhere to the boundaries specified by the lower and upper limits of the solution search space during the initialization phase. At the start of the Differential Evolution process, the generation of $x_{jG}$ occurs as follows:

$$x_{jG} = x_{lower} + (x_{upper} - x_{lower}) \times rand[0,1] \quad (3)$$

Here, $x_{lower}$ and $x_{upper}$ represent the lower and upper bounds of the search space, respectively. $rand[0,1]$ is a probability distribution function that generates values within the specified range [0,1] with equal probability.

### 3.6.2.2 Mutation

Mutation explores new regions of the solution space beyond the current population. It promotes diversity by introducing random perturbations. This helps in finding better solutions during the optimization process. It generates a mutation vector by randomly combining different vectors in accordance with following equation:

$$VT_{j,G} = x_{BestG} + F \times (x_{r_1G} - x_{r_2G}) \quad (4)$$

Where $VT_{j,G}$ is mutant vector, and $x_{BestG}$ signifies the best solution found up to generation G. Additionally, $x_{r_1G}$ and $x_{r_2G}$ are random base vectors (where $r_1 \neq r_2 \neq Best_G$ and $Best_G, r_1, r_2 \in \{1, 2, \ldots, NP\}$). F, serving as the scaling factor, regulates the mutation process, with its value constrained within the range of [0, 1].

### 3.6.2.3 Crossover

In the crossover stage the mutated vector $VT_{j,G} = \{VT_{1,j,G}, VVT_{2,j,G} \ldots, VT_{D,j,G}\}$ and the target vector $x_{j,G} = \{x_{1,j,G}, x_{2,j,G} \ldots x_{D,j,G}\}$ are compared component-wise using a Crossover Probability or Crossover Rate (CPr), which ranges between 0 and 1, to determine the selection of components for the resulting vector in the crossover stage. If a component of the mutant vector is selected, it is included in the crossover result vector, referred to as the trial vector denoted as $XX_{j,i,G}$. Otherwise, the corresponding component from the target vector is chosen. This process can be represented as follows:

$$XX_{j,i,G} = \begin{cases} VT_{i,j,G} & \text{if } j \epsilon CrossPoints \\ x_{i,j,G} & \text{otherwise} \end{cases} \quad (5)$$

In equation 5, CrossPoints signifies a collection of crossover points established through consideration of both the CPr value and the selected crossover method, which, in our study, is Binomial. To ensure that CrossPoints is non-empty, a random member $j_{rand} \epsilon \{1,2,\ldots,D\}$ is chosen. Subsequently, a new member, denoted as m $\epsilon \{1,2,\ldots,D\}$, is included in CrossPoints under the following condition:

$$CrossPoints = CrossPoints \cup \{m\} \times \delta\{rand[0,1] < CPr \wedge mj_{rand}\}, \text{ for } j \epsilon \{1,2,\ldots,D\} \quad (6)$$

Here, δ is a Kronecker delta function used to determine whether the addition of the new element m should occur.

### 3.6.2.4 Selection

In the selection stage, the fitness values of the target vector $x_{jG}$ and the crossover result vector $XX_{jG}$ are compared. After comparing $x_{jG}$ and $XX_{jG}$, the vector with lower fitness value is chosen to be promoted to the next generation. This process can be illustrated in the following manner:

$$X_{j,G+1} = \begin{cases} XX_{jG} & \text{if } f(XX_{jG}) \leq f(x_{jG}) \\ x_{jG} & \text{otherwise} \end{cases} \quad (7)$$

In equation 7, $f(XX_{jG})$ denotes the process of evaluating the objective function for the crossover result vector, while $f(x_{jG})$ represents the process of evaluating the objective function for the target vector. The comparison $f(XX_{jG}) \leq f(x_{jG})$ then represents the evaluation of whether the objective function value for the crossover result vector is less than or equal to the objective function value for the target vector. The result of this comparison determines which vector should be included in the next generation's population.

In our categorical task of classifying COVID-19 from CXR images, we use the loss function of our proposed ensemble as the fitness criterion to measure solution quality during the optimization of fuzzy densities. The ensemble loss, derived from the Categorical Cross Entropy function, evaluates the difference between predicted probabilities and actual labels assigned to the image classes. This loss serves as a measure of the solution's effectiveness provided by the Differential Evolution algorithm, reflecting the disparity between predicted and actual outcomes. We use the ensemble model's loss on the validation data as a fitness metric for evaluating candidate solutions. This ensures impartial assessment, as it reflects the model's real performance on unseen data by relying solely on validation data, without incorporating test data.

The mutation, crossover, and selection steps in Differential Evolution are repeated until a termination criterion is met. The termination criterion defines when the algorithm should stop. In this study, we set a maximum of 100 generations as the termination criterion. Once this condition is met and fuzzy densities are determined, we proceed with aggregating features from the three dense layers by applying the Choquet integral operator to these vectors.

### 3.6.3 Fuzzy Choquet Integral Layer

The aggregating process using the Choquet function involves considering a real-valued function, h:X→R, which represents the evidence or support for a hypothesis (such as features or output votes from DCNNs). The specific relation for obtaining the aggregated feature vector using the Choquet operator can be represented by the following equation:

$$C_g(h, g_\Omega) = \sum_{i=1}^{N}(h(x_{\Pi(i)})(g_\Omega(A_i) - g_\Omega(A_{i-1}))) \qquad (8)$$

The function "h" maps real values to the elements within set X. Here, these values are specifically obtained from the outputs of dense layers, each containing 256 neurons activated by ReLU activation function, in the three DCNNs (Criteria). Given that "h" produces real numbers, the outputs from the three Dense layers can act as supporting hypotheses for each input image class. These supporting hypotheses are consolidated using the Choquet operator, resulting in an aggregated vector. This aggregated vector is then used to assign each input image to its respective class, with the Choquet output directed to the prediction layer of DenseNet-201.

In equation 8, Each $\pi$ signifies a permutation of X, where function "h" applied to permuted entries satisfies $h(x_{\Pi(1)}) > h(x_{\Pi(2)}) > \cdots > h(x_{\Pi(N)})$. The notation $g_\Omega(A_i)$, with $A_i = \{x_{\Pi(1)}, x_{\Pi(2)} \dots, x_{\Pi(N)}\}$, represents fuzzy measures associated with these permutations. In this study, N, which is 3, corresponds to the number of DCNNs. The individual elements from the outputs of the three dense layers undergo pairwise permutation. Elements at corresponding positions across the vectors are compared and arranged in descending order. We have provided the Pseudo-Code of our proposed Choquet layer in **Fig. 2** further portrays the operation of our proposed Choquet layer applied to three feature vectors originating from three DCNNs, resulting in an aggregated vector. In **Fig. 2**, the Choquet layer manages three vectors: FD representing the feature vector of DenseNet-201, FI representing the feature vector of Inception-V3, and Fx representing the feature vector of Xception. The resulting vector of aggression is denoted as FCH, representing the feature vector of the Choquet layer.

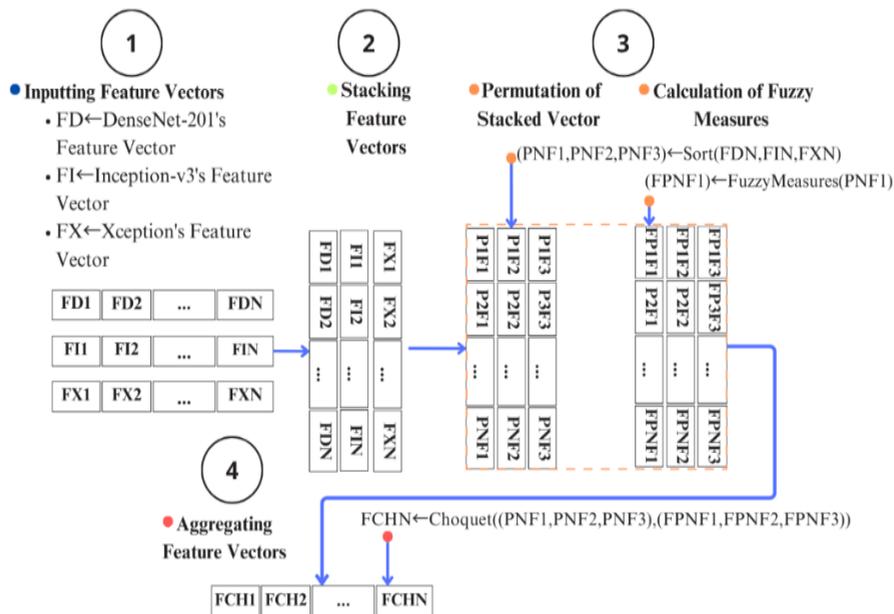

**Fig. 2** Visualization of the Choquet layer operation applied to three vectors

**Algorithm 1** Pseudo-Code of Choquet layer

**Algorithm 1: Psuedo-Algorithm of Choquet layer**

**Input :**
  $X_{DCNNs}$: A tensor of supporting hypotheses derived from the last hidden layers of DCNNs.
  $F_M$: A tensor containing fuzzy Densities.
  $\lambda$: A $\lambda$-value-containing tensor.

**Output :**
  $SH_{Choquet}$: An aggregated tensor of supporting hypotheses.

1:   $SH_{Choquet} \leftarrow$ Empty list
2:   $X_{St} \leftarrow$ Stack tensors in $X_{DCNNs}$ along the third axis using TensorFlow's stack function.
3:   $X_{St_{sorted}} \leftarrow$ Sort tensor $X_{St}$ along its last axis in descending order to perform permutation using TensorFlow's sort function.
4:   $F_{M_{sorted}} \leftarrow$ Fetch elements from $F_M$ based on the reverse of the sorted indices of $X_{St}$ along the third axis using TensorFlow's gather function.
5:   $F_{M_{First}} \leftarrow$ Fetch elements from $F_{M_{sorted}}$ at index 0 along axis 2.
6:   $X_{St_{sorted_{First}}} \leftarrow$ Fetch elements from $X_{St_{sorted}}$ at index 0 along axis 2.
7:   $SH_{Choquet_{First}} \leftarrow X_{St_{sorted_{First}}} \times F_{M_{first}}$.
8:   Append $SH_{Choquet_{First}}$ to $SH_{Choquet}$.
9:   $F_{M_N} \leftarrow F_{M_{First}}$.
10:   for $idx$ in range(1, length($X_{DCNNs}$)) do:
11:     $F_{M_{idx}} \leftarrow$ Fetch elements from $F_{M_{Sorted}}$ using the indices provided by $idx$ along axis 2.
12:     $Fuzzy_{\lambda_{idx}} \leftarrow F_{M_N} + F_{M_{idx}} + (F_{M_{idx}} \times F_{M_N} \times \lambda)$ #Calculate $Fuzzy_{\lambda_{idx}}$ according to Sugeno-$\lambda$ measures theory.
13:     $X_{St_{sorted_{idx}}} \leftarrow$ Fetch elements from $X_{St_{sorted}}$ using the indices provided by idx along axis 2.
14:     $SH_{Choquet_{idx}} \leftarrow X_{St_{sorted_{idx}}} \times (Fuzzy_{\lambda_{idx}} - F_{M_N})$
15:     Append $SH_{Choquet_{idx}}$ to $SH_{Choquet}$.
16:     $F_{M_N} \leftarrow Fuzzy_{\lambda_{idx}}$.
17:   $F_{M_{Last}} \leftarrow$ Create a constant tensor with a value of 1 #Unity at entire set ($g_\Omega(X) = 1$).
18:   $X_{St_{sorted_{Last}}} \leftarrow$ Fetch elements from $X_{St_{sorted}}$ at index corresponding to the length of $X_{DCNNs}$ along axis 2.
19:   $SH_{Choquet_{Last}} \leftarrow X_{St_{sorted_{Last}}} \times (F_{M_{Last}} - F_{M_N})$.
20:   Append $SH_{Choquet_{Last}}$ to $SH_{Choquet}$.
21:   $SH_{Choquet} \leftarrow$ Calculate the sum of elements in $SH_{Choquet}$ along axis 2 using TensorFlow's reduce_sum function.

## 4. Experimental Results

### 4.1. Dataset Description

We used the COVIDx benchmark dataset [24] for both training and testing phases. We opted for COVIDx versions 8A/B instead of 9A/B, given our resource constraints and reliance on transfer learning with DCNNs pre-trained on ImageNet. This choice reflects the suitability of these versions for our limited computational resources. The COVIDx dataset has two variations. COVIDxA includes three categories: Normal, COVID-19, and Pneumonia. COVIDxB is binary, containing COVID-19 and non-COVID-19 cases. Researchers can compare their model performance with ours using either version—COVIDx8 or COVIDx9—since both share similar test samples.

We have implemented the train-test split according to the authors' provided labels, ensuring that the sizes of our training and testing sets, as well as the images in both the training and testing sets, match those available in the COVIDx8 train set and test set. Moreover, we created a validation dataset by applying a 90/10 split rate for training/validation from the original training set, and the test set was not split, as the COVIDx repository provides its own test set.

COVIDx8A includes 1215 COVID-19, 7966 Normal, and 5475 Pneumonia images. The test set consists of 200 COVID-19 images and 100 images each for Normal and Pneumonia classes. Similarly, COVIDx8B contains an equivalent number of images as COVIDx8A. In COVIDx8B, the none-COVID19 category comprises images obtained by combining those from the Pneumonia and Normal classes, while the COVID-19 category in COVIDx8B aligns exactly with the COVID-19 images in COVIDx8A.

**Fig. 3** illustrates the data distribution across the three classes in the training and testing sets of the COVIDx8 Dataset. **Fig. 4** also displays a compilation of images from the COVIDx dataset, accompanied by their respective labels.

### 4.2. Experimental Setup

To implement our proposed approach, we selected Python as our programming language and utilized TensorFlow along with its Keras package. Our code was executed in the Google Colab environment, which provided the following hardware resources at no cost: Nvidia Tesla T4 GPU with 15.6 GB memory, 12.6 GB RAM, Intel Xeon CPU @ 2.20 GHz, and 107.7 GB HDD.

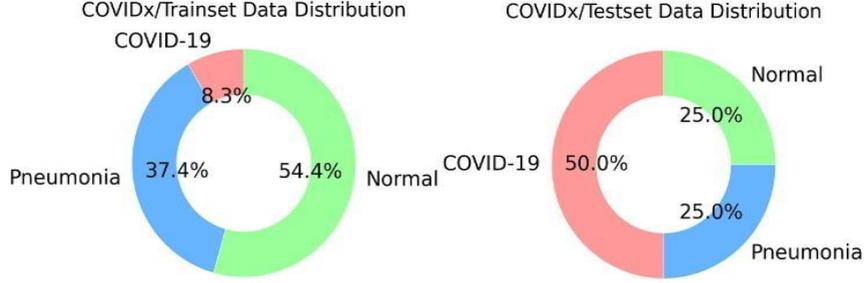
**Fig. 3** The distribution of the three classes in the train and test sets of the COVIDx8 dataset.

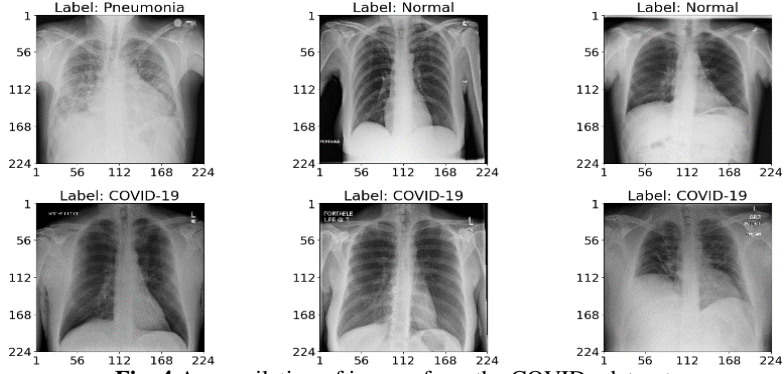
**Fig. 4** A compilation of images from the COVIDx dataset.

## 4.3. Evaluation Metrics

Before defining the evaluation metrics used in this study, it's crucial to clarify the meanings of True Positives (TP), True Negatives (TN), False Positives (FP), and False Negatives (FN), particularly about the COVID-19 class: TP signifies individuals within the COVID-19 class, confirmed to have the virus, correctly identified as positive by the model. FP refers to cases where individuals outside the COVID-19 class are mistakenly identified as positive. TN represents individuals not belonging to the COVID-19 class, correctly identified as negative by the model, while FN indicates individuals within the COVID-19 class incorrectly identified as negative. Based on TP, FP, TN, and FN, the evaluation metrics utilized in this study to assess the performance of the proposed ensemble model and DCNNs are as follows:

$$Accuracy = \frac{TP+TN}{TP+FP+TN+FN} \quad (9)$$

$$Precision = \frac{TP}{TP+FP} \quad (10)$$

$$\frac{Recall}{Sensitivity} = \frac{TP}{TP+FN} \quad (11)$$

$$F1-score = \frac{2 \times Precision \times Recall}{Precision + Recall} \quad (12)$$

$$Area\ Under\ the\ ROC\ Curve\ (AUC) = \frac{\left(\frac{TP}{TP+FN}+\frac{TN}{TN+FP}\right)}{2} \quad (13)$$

$$Specificity = \frac{TN}{TN+FP} \quad (14)$$

$$Matthews\ correlation\ coefficient\ (MCC) = \frac{TP \times TN - FN \times FP}{\sqrt{(TP+FN) \times (TN+FP) \times (TP+FP) \times (TN+FN)}} \quad (15)$$

In the AUC curve, the vertical axis represents the True Positive Rate, also known as sensitivity. The horizontal axis represents the False Positive Rate. The False Positive Rate is calculated as FP/(TN + FP). The top-left corner of the AUC graph is often seen as the ideal position. It represents a False Positive Rate of zero and a True Positive Rate of one. However, this scenario may not accurately reflect real-world situations.

## 5. Evaluation
### 5.1. Models' Performance
In this study, we employed DenseNet-201, Inception-V3, and Xception as individual DCNNs for our ensemble model. This section presents the performance outcomes of these DCNNs and the proposed ensemble model, along with a comparative analysis.

**Table 2** details the fuzzy density, validation accuracy, and λ coefficient for each DCNN. Additionally, **Fig. 5** illustrates the fluctuations in training and validation accuracy and error across epochs for all three models. Based on **Table 2** and **Fig. 5**, DenseNet-201 demonstrated superior validation accuracy and lower Categorical Cross-Entropy loss compared to Inception-V3 and Xception. However, the Differential Evolution optimization assigned higher fuzzy density values to Inception-V3 and Xception.

**Table 1** highlights the ensemble model's performance on the COVIDx8A test set, achieving a precision of 100% for COVID-19 detection. This indicates no false positives and perfect identification of true positives. The recall rate of 99% reflects the model's ability to detect most COVID-19 cases with minimal false negatives, making it a reliable alternative to RT-PCR tests. Additionally, the model achieved 100% specificity, with MCC, AUC, and F1-score values of 99%, 99.59%, and 99.49%, respectively.

For the Normal class, the model recorded 96.08% precision, 98% recall, 98.66% specificity, 96.03% MCC, 98.33% AUC, and a 97.03% F1-score. For Pneumonia, it achieved 96% precision and recall, 98.66% specificity, 94.66% MCC, 97.33% AUC, and a 96% F1-score. Despite COVID-19 being a minority class in the dataset, the ensemble model achieved top performance across all metrics for its detection.

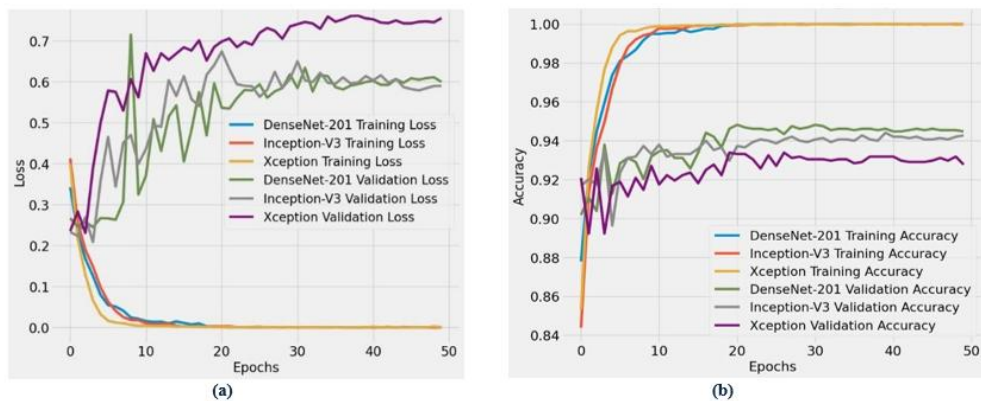

**Fig. 5** Learning curves during training for (a) loss, (b) accuracy

**Table 1** valuation results for the proposed ensemble model

| Metric (%) | COVID-19 | Normal | Pneumonia |
| --- | --- | --- | --- |
| **Precision** | 100.00 | 96.078 | 96.00 |
| **Recall** | 99.00 | 98.00 | 96.00 |
| **Specificity** | 100.00 | 98.66 | 98.66 |
| **MCC** | 99 | 96.03 | 94.66 |
| **AUC** | 99.50 | 98.33 | 97.33 |
| **F1-Score** | 99.49 | 97.03 | 96.00 |

To explain why Differential Evolution assigned higher densities to Inception-V3 and Xception despite their lower validation accuracies, it's important to note that we used the Choquet integral to handle synergy and redundancy, which are complex concepts. Redundancy involves having multiple elements that provide similar or overlapping details. Synergy, on the other hand, refers to the interaction or fusion of different components, resulting in insights or information that go beyond what each individual component can offer alone. Based on this understanding, we used Differential Evolution to adjust the densities for the DCNNs in the ensemble model. We treated them as a single set, not as separate units with their own importance. With this, we can consider the impact of individual singletons and their interactions on the quality of the solution. Because of this approach, Differential Evolution did not give higher density to DenseNet-201, a singleton, just because of its higher validation accuracy. We describe

Differential Evolution as performing the optimization process collectively because it evaluates both the crossover result and the target vector quality using the ensemble loss. This loss is influenced by the membership values of individual DCNNs and their subsets.

Table 2 fuzzy density, validation accuracy, and validation loss for all Models and $\lambda$ value

| | |
|---|---|
| 0.12470619 | **Fuzzy density of DenseNet-201** |
| 0.29971752 | **Fuzzy density of Inception-V3** |
| 0.2989895 | **Fuzzy density of Xception** |
| 0.9482 | **Dense-Net-201 validation accuracy** |
| 0.9441 | **Inception -V3 validation accuracy** |
| 0.9338 | **Xception validation accuracy** |
| 0.5366 | **Dense-Net-201 validation loss** |
| 0.6137 | **Inception -V3 validation loss** |
| 0.6848 | **Xception validation loss** |
| 0.9509 | **Ensemble model validation accuracy** |
| 0.1531 | **Ensemble model validation accuracy** |
| 1.5253944 | $\lambda$ **Value** |

In **Fig. 6**, we provided the confusion matrix of the proposed ensemble model for classifying COVIDx8A instances. Examining **Fig. 6**, 198 instances of COVID-19 were correctly classified as COVID-19, resulting in 198 TP cases. At the same time, 200 instances were correctly identified as not COVID-19, resulting in 200 TN cases. The proposed ensemble model misclassified only 2 COVID-19 cases as Pneumonia, despite the dataset's imbalance, with more COVID-19 images in the test set and fewer in the training set. This resulted in 2 FN instances for the COVID-19 class. Remarkably, there are no occurrences of FP, as no instances of Normal or Pneumonia were predicted to be infected with COVID-19, resulting in a Precision of 100% for this class. For the Normal class, the count of TP stands at 98, with 296 TN, 4 FP, and 2 FN. Regarding the Pneumonia class, there are 96 instances of TP, 296 instances of TN, along with 4 instances of FP, and 4 instances of FN.

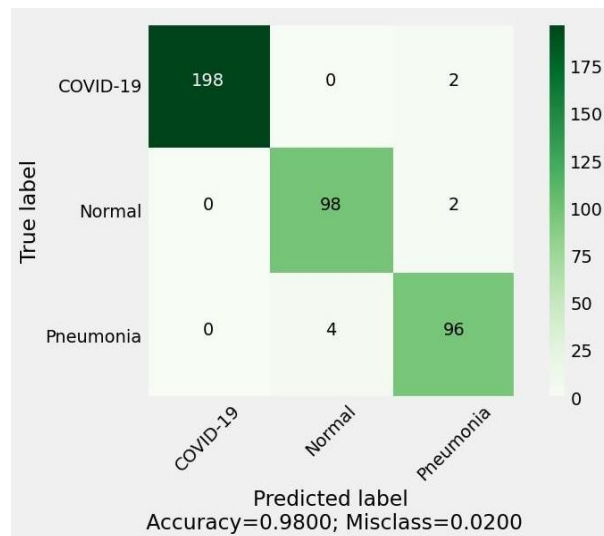

**Fig. 6** Confusion matrix for the proposed ensemble model

Moreover, we conducted a comparative analysis of the models by evaluating their ROC curves. **Fig. 7** presents the ROC curves for all models, while exclusively showcasing the multi-labeled ROC curve for the proposed ensemble model. In **Fig. 7**, the top-left point represents the optimal point with a False Positive Rate of zero and a True Positive Rate of one. The proposed ensemble model is closer to this optimal point compared to individual DCNNs, achieving a lower False Positive Rate and a higher True Positive Rate. Additionally, the ROC curve of the ensemble model highlights its superior performance in identifying COVID-19, with a higher True Positive Rate and lower False Positive Rate for the COVID-19 class. This advantage is due to the precise alignment of the COVID-19 ROC curve with the optimal point.

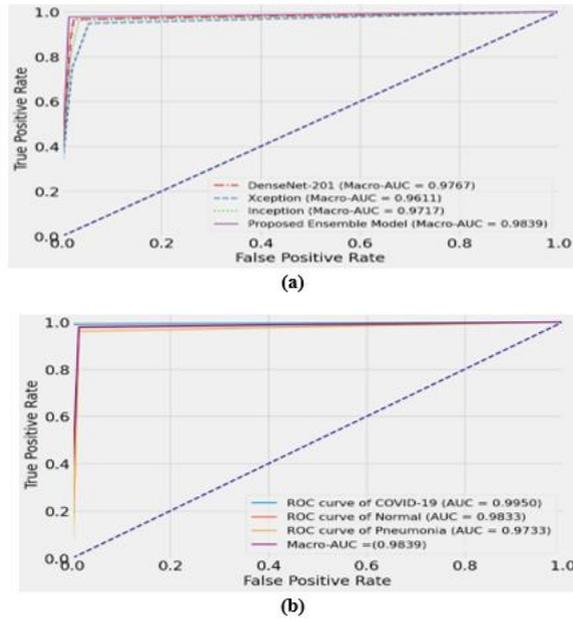

**Fig. 7** ROC Curves for (a) all models, (b) the ensemble model

**Table 3** presents a comprehensive overview of the models' performance across accuracy, precision, recall, specificity, MCC, AUC, and F1-Score. The information presented in **Table 3** helps highlight the overall effectiveness and distinctive strengths of each model in classifying COVIDx8A test set images. However, it does not provide performance insights for individual classes.

Per the findings in **Table 3**, the proposed ensemble model attained the highest precision, reaching a rate of 97.35%. This result suggests that among all the individual DCNNs, the ensemble model is expected to have had the lowest occurrence of false positive predictions. Turning to individual DCNNs, DenseNet-201 displayed the strongest precision among Xception and Inception-V3, achieving a precision of 96.08%. Following DenseNet-201, Inception-V3 exhibited a precision of 95.18%, while Xception achieved a precision of 93.21%.

In terms of accuracy, the proposed ensemble model demonstrated its superiority with the highest rate at 98.00%, showcasing its enhanced ability over its components to accurately predict both positive and negative instances.

**Table 3** accuracy, precision, recall, specificity, MCC, AUC, and F1-score values for DCNNs and the proposed ensemble model in three-class classification

| Metric (%) | DenseNet-201 | Inception-V3 | Xception | Proposed ensemble model |
|---|---|---|---|---|
| **Accuracy** | 97.00 | 96.25 | 94.50 | 98.00 |
| **Precision** | 96.08 | 95.18 | 93.21 | 97.35 |
| **Recall** | 96.66 | 96.00 | 94.66 | 97.67 |
| **Specificity** | 98.66 | 98.33 | 97.55 | 99.11 |
| **MCC** | 95.23 | 94.07 | 91.14 | 96.81 |
| **AUC** | 97.67 | 97.17 | 96.11 | 98.39 |
| **F1-Score** | 96.36 | 95.56 | 93.82 | 97.50 |

The reported accuracy of 97.00% achieved by DenseNet-201 during the evaluation phase outperformed other DCNNs, with Inception-V3 achieving 96.25% and Xception reaching 94.50%. The proposed ensemble model demonstrated the highest recall rate at 97.67% (Table 3), effectively minimizing false negatives (FN) and accurately detecting positive instances. DenseNet-201 attained a recall of 96.66%, exceeding Inception-V3 (96%) and Xception (94.66%) but falling 1% short of the ensemble model.

In terms of specificity, the ensemble model achieved the highest score of 99.11%, reflecting its exceptional ability to avoid false positives (FP) and accurately classify negative cases. DenseNet-201 followed with 98.66%, while Inception-V3 and Xception achieved 98.33% and 97.55%, respectively. This indicates that the ensemble model's negative predictions are more reliable compared to the individual DCNNs.

The ensemble model also recorded the highest MCC at 96.81%, showcasing its balanced performance across true positives (TP), true negatives (TN), FP, and FN. DenseNet-201 followed with 95.23%, while Inception-V3 and Xception achieved 94.07% and 91.14%, respectively. Similarly, the ensemble model achieved the highest F1-Score at 97.50%, highlighting its superior balance between precision and recall. It also had the highest AUC of 98.39%. Among the individual DCNNs, DenseNet-201 outperformed the others with an F1-Score of 96.36% and an AUC of 97.67%, while Inception-V3 recorded 95.56% and 97.17%, and Xception scored 93.82% and 96.11%. The use of the Choquet layer for feature vector aggregation and Differential Evolution for determining fuzzy densities enabled the ensemble model to outperform the individual DCNNs across all metrics.

To further assess robustness, the DCNNs and ensemble models were fine-tuned using the COVIDx Binary dataset (COVIDx8B). As shown in **Table 4**, the ensemble model achieved the highest accuracy, precision, recall, F1-Score, MCC, and AUC, mirroring its performance on the COVIDx8A dataset. This consistent superiority in both 3-class and 2-class scenarios highlights the model's ability to generalize, reduce overfitting, and integrate the strengths of the three DCNNs under different conditions.

The computational efficiency of the Choquet layer is detailed in **Table 5**, where it aggregates evidence vectors of varying dimensions based on three criteria. For inputs such as (3, 1,000,000, 3), the aggregation process took only 7.5338 seconds using 0.4 GB of GPU memory. Similarly, for three feature vectors with dimensions of (100,000, 256), the aggregation was completed in 61.6955 seconds with 0.6 GB of GPU usage. These results demonstrate the scalability of the Choquet layer, which effectively processes large datasets while maintaining efficient time and resource usage. This scalability ensures real-world applicability, allowing valuable insights to be derived from extensive data within practical timeframes.

**Table 4** accuracy, precision, recall, specificity, MCC, AUC, and F1-score values for DCNNs and the proposed ensemble model in binary classification

| Metric (%) | Dense Net-201 | Inception-V3 | Xception | Proposed Model |
|---|---|---|---|---|
| **Accuracy** | 98.75 | 98.50 | 98.00 | **99.50** |
| **Precision** | 98.78 | 98.54 | 98.07 | **99.505** |
| **Recall** | 98.75 | 98.50 | 98.00 | **99.50** |
| **Specificity** | 98.75 | 98.50 | 98.00 | **99.50** |
| **MCC** | 97.53 | 97.04 | 96.07 | **99.00** |
| **AUC** | 98.75 | 98.50 | 98.00 | **99.50** |
| **F1-Score** | 98.75 | 98.50 | 97.99 | **99.50** |

**Table 5** performance of the Choquet layer processing across varying input vector dimensions

| Dimensions of input vectors | Time (GPU) | GPU usage | Batch size |
|---|---|---|---|
| (3,100000,256) | 61.6955s | 0.6 GB | 50,000 |
| (3,1000000,3) | 7.5338s | 0.4 GB | 50,000 |

# 6. Comparison with Other Methods

Conducting precise comparisons between our proposed ensemble model and other existing methods for automatic COVID-19 identification from CXR images using deep learning-based frameworks presents challenges. This is due to the diversity of datasets and variations in data versions used by authors in the literature [23]. To maintain scientific rigor, we limit our comparison of the proposed ensemble model to similar studies in the literature that also used the COVIDx Dataset, similar to our approach. In

**Table 6**, we presented the performance of our proposed ensemble models alongside those from prior studies that utilized the COVIDx dataset, with all results based on this dataset. Upon examination of

**Table 6**, it's evident that our proposed ensemble model outperforms all models from the papers [22], [23], [24], [36], [43], and [46] across both three-class and two-class classifications.

# 7. Conclusion and Future Directions

In our study, we introduced a novel ensemble model that incorporates feature vector aggregation using a non-linear ensembling operator known as the fuzzy Choquet integral. The ensemble model we proposed integrates the last hidden layers of three pre-trained DCNNs: DenseNet201, Inception-V3, and Xception. For this purpose, we designed our newly developed Choquet layer, specifically tailored for efficient information aggregation using the Choquet integral method. To implement this layer, we utilized the TensorFlow framework for its inherent parallel processing capabilities, as well as for the flexibility and scalability it offers. In our approach to calculate Choquet function parameters (fuzzy measures), which are used to model interactions among DCNNs, we utilized the Sugeno-$\lambda$ measure theory and employed the Differential Evolution algorithm. With the Differential Evolution algorithm, we determined the best-fit fuzzy membership value of each DCNN, and using Sugeno-$\lambda$ measure theory and the fuzzy membership values, we computed fuzzy measures for subsets of DCNNs.

Table 6 comparing our proposed ensemble model with previous methods

| Reference | Dataset | Accuracy | Precision | Recall | Specificity | MCC | AUC | F1-Score |
|---|---|---|---|---|---|---|---|---|
| Proposed Model | COVIDxA | **98.00** | **97.35** | **97.67** | **99.11** | **96.81** | **98.39** | **97.50** |
| | COVIDxB | **99.50** | **99.505** | **99.50** | **99.50** | **99.00** | **99.50** | **99.50** |
| Banerjee et al. [22] | COVIDxA | 96.37 | 95.69 | 96.30 | 97.52 | 93.31 | - | 96.30 |
| | COVIDxB | 99.49 | 99.23 | 96.46 | 96.46 | 95.66 | 96.46 | 97.80 |
| Bhowal et al. [23] | COVIDxA | 93.81 | - | - | - | - | - | - |
| Wang et al. [24] | COVIDxA | 93.30 | - | - | - | - | - | - |
| Nayak et al. [46] | COVIDxA | 96.57 | 93.51 | - | - | - | - | 93.50 |
| Ullah et al. [43] | COVIDxB | 99.62 | - | 91.07 | 99.62 | - | - | - |
| Ucar et al. [36] | Augmented COVIDxA | 98.26 | 98.26 | 98.26 | 99.13 | 97.39 | | 98.25 |
| | COVIDxA | 76.37 | 75.26 | 69.21 | 79.93 | 55.27 | - | 66.89 |

In our future research pursuits, we aim to integrate CT scan image repositories and fine-tune pre-trained DCNNs on CT images to optimize our ensemble model for this modality. We will either use DCNN weights trained on the COVIDx dataset as the starting point or fine-tune them with ImageNet pre-trained weights on CT image datasets. This approach will improve the model's effectiveness for broader and more practical applications.

In this study, our primary focus was on aggregating the outputs of the last hidden layers of DCNNs, and we did not train the ensemble model as a single entity. For future research, the prospect of training our proposed ensemble model as a unified entity or adjusting specific parameters within the ensemble model holds promise for enhancing its performance. Given that the core ensembling mechanism of the proposed model is the Choquet function, the updating parameters of models during optimization will be influenced by the functionality of the Choquet integral. Therefore, the optimization process, unlike separately fine-tuning DCNNs, fosters interactions between models, enabling them to learn from each other and adapt their contributions to the ensemble's decision-making process.

Furthermore, putting into practice a diverse range of metaheuristic optimization algorithms, including PSO, GAs, Teaching-Learning-Based Optimization (TLBO), Simulated Annealing (SA), Ant Colony Optimization (ACO), Harmony Search (HS), and Artificial Bee Colony (ABC), to compute fuzzy densities, or even more comprehensively, tapping into them to calculate fuzzy measures for subsets of DCNNs, can greatly aid in determining the most effective approach for enhancing the ensemble model's performance through the concept of fuzzy measures.